\documentclass[sigconf, letter]{acmart}

\usepackage{booktabs}

\usepackage{flushend}
\usepackage{graphicx}
\usepackage{cleveref}
\usepackage[caption=false,font=footnotesize]{subfig}
\graphicspath{ {Images/} }

\setcopyright{acmcopyright}
\copyrightyear{2018}
\acmYear{2018}
\setcopyright{acmcopyright}
\acmConference[GECCO '18]{Genetic and Evolutionary Computation Conference}{July 15--19, 2018}{Kyoto, Japan} \acmBooktitle{GECCO '18: Genetic and Evolutionary Computation Conference, July 15--19, 2018, Kyoto, Japan}
\acmPrice{15.00} \acmDOI{10.1145/3205455.3205541} \acmISBN{978-1-4503-5618-3/18/07}

\begin{document}
\title{Towards the Targeted Environment-Specific Evolution of Robot Components}

\author{Jack Collins}
\affiliation{%
  \institution{CSIRO \& Queensland University of Technology}
  \city{Brisbane}
  \state{Queensland}
}
\email{jack.collins@csiro.au}

\author{Wade Geles}
\affiliation{%
  \institution{CSIRO \& University of Queensland}
  \city{Brisbane}
  \state{Queensland}
}
\email{wade.geles@csiro.au}

\author{David Howard}
\affiliation{%
  \institution{CSIRO}
  \streetaddress{1 Technology Court, Pullenvale}
  \city{Brisbane}
  \state{Queensland}
  \postcode{4068}
}
\email{david.howard@csiro.au}

\author{Frederic Maire}
\affiliation{
  \institution{Queensland Univeristy of Technology}
  \streetaddress{2 George Street, Brisbane City}
  \city{Brisbane}
  \state{Queensland}
  \postcode{4000}
}
\email{f.maire@qut.edu.au}

\renewcommand{\shortauthors}{J.Collins, et al.}

\begin{abstract}
This research considers the task of evolving the physical structure of a robot to enhance its performance in various environments, which is a significant problem in the field of Evolutionary Robotics.  Inspired by the fields of evolutionary art and sculpture, we evolve only targeted parts of a robot, which simplifies the optimisation problem compared to traditional approaches that must simultaneously evolve both (actuated) body and brain.  Exploration fidelity is emphasised in areas of the robot most likely to benefit from shape optimisation, whilst exploiting existing robot structure and control. Our approach uses a Genetic Algorithm to optimise collections of Bezier splines that together define the shape of a legged robot's tibia, and leg performance is evaluated in parallel in a high-fidelity simulator.  The leg is represented in the simulator as 3D-printable file, and as such can be readily instantiated in reality.  Provisional experiments in three distinct environments show the evolution of environment-specific leg structures that are both high-performing and notably different to those evolved in the other environments.  This proof-of-concept represents an important step towards the environment-dependent optimisation of performance-critical components for a range of ubiquitous, standard, and already-capable robots that can carry out a wide variety of tasks.
\end{abstract}




\maketitle

\section{Introduction}

Robot design typically involves an extensive degree of expert knowledge, a correspondingly large amount of man-hours, and a constrained number of iterations with abstract, high-level constraints in place of potentially complex environmental interactions.  As design decisions are influenced by human preconceptions, and because humans do not deal well with high-dimensional search spaces, the design space considered is typically artificially limited.  This is especially true given (i) the increasing ubiquity of additive manufacturing, allowing us to physically create increasingly free-form structures, and (ii) a rapidly-expanding assortment of construction materials for those structures. These advances increase the attainable design space of robots beyond levels that are tractable in a traditional engineering approach.

An attractive alternative involves creating algorithms that in turn create environmentally-specialized robots.  Evolutionary Robotics (ER)~\cite{EvolutionaryRoboticsTheBiology} creates bespoke robots by iteratively searching in a space of possible bodies and/or controllers using an evolutionary algorithm.  Evolutionary algorithms handle high-dimensional design spaces well~\cite{eiben2003introduction}, are bias-free in terms of the solutions generated, and permit the generation of novel and surprising solutions which may not be considered in other approaches.  In some cases, evolved solutions are capable of outperforming the best traditionally-engineered approaches~\cite{7559707}.  

\begin{figure}[t!]
      \centering
      \includegraphics[width=8cm]{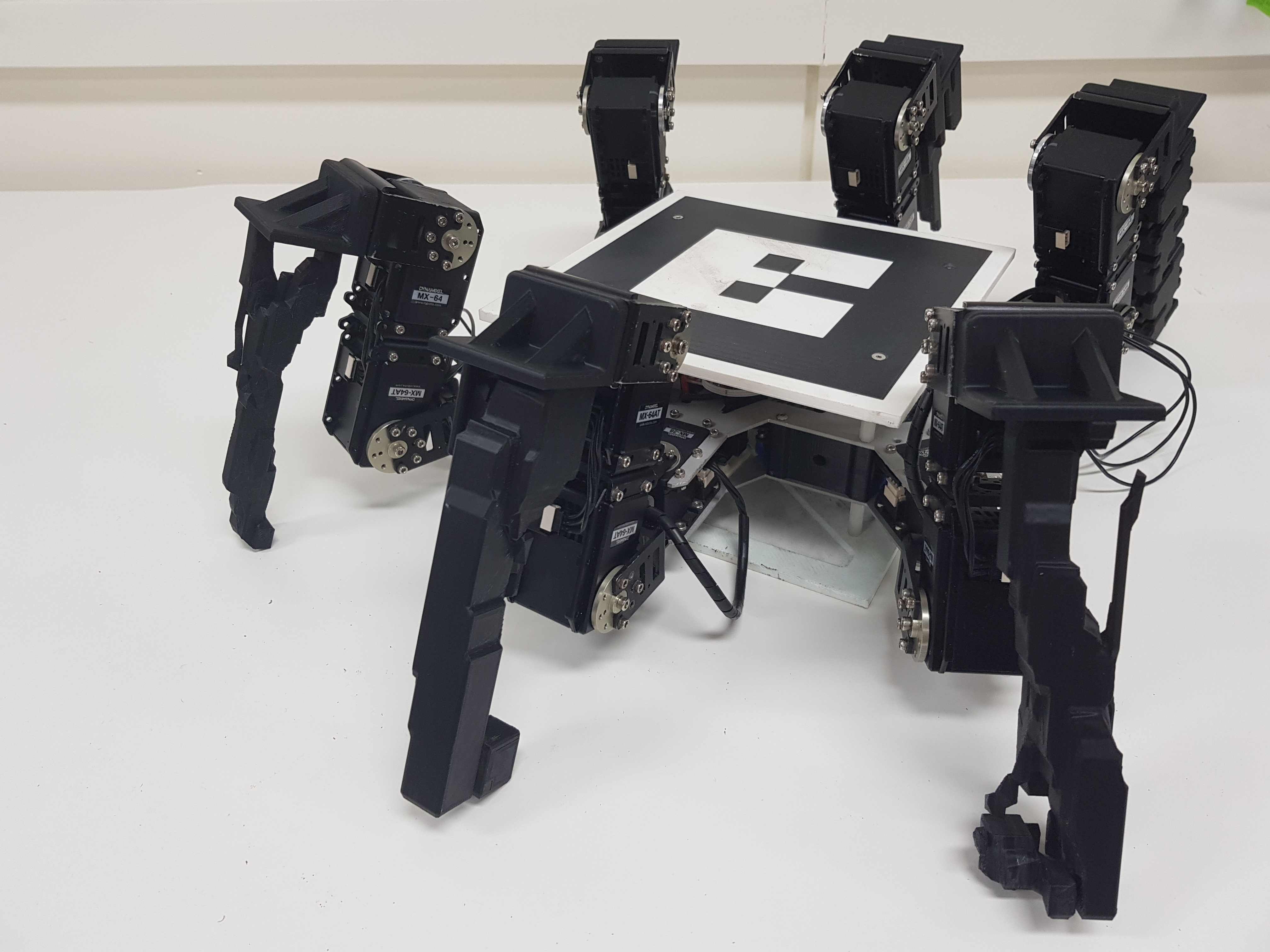}
      \caption{Exemplar hexapod legs generated through evolution, printed together with their mounting brackets and attached to a CSIRO hexapod.  Brackets are identical for each leg, and, after inserting four metal threads into pre-printed holes, attach to a leg actuator using 3mm bolts.  Different sets of legs may be swapped onto the hexapod depending on the prevalent environment type in its current mission.}
      \label{real_legs}
  \end{figure}

In contrast to co-evolutionary approaches that adapt body and brain together,\footnote{To date, resulting in solutions that, although 'complete', lack real-world applicability.} we focus on specialising only the most performance-critical parts of an already-established morphology/controller combination, in this case evolving only the leg of a hexapod robot.  Figure \ref{real_legs} shows such a robot with a variety of evolved legs attached.  This approach allows us to improve environmental performance of an already useful and functional platform, without having to evolve an entire robot or controller from scratch.  Our approach therefore combines engineered and evolved; creating a trade-off that constrains attainable morphologies, in favour of solutions that are fully-functioning, useful, and applicable to a wide variety of tasks.\footnote{Hexapods, for example, are used for missions including confined space inspection, navigation in uneven terrain, and disaster response.}

Ongoing advances in additive manufacture make real morphology design an increasingly active research problem in ER.  As part of this push towards evolving real, physical artefacts --- dubbed the {\em Evolution of Things}~\cite{FromEvolutionaryComputation}, our components are imported into a high-fidelity simulator as 3D printable files.  As such, they can be easily instantiated in the real world.  

Components are represented by collections of Bezier splines, which collectively define the 3D morphology of the leg. A Genetic Algorithm (GA)~\cite{holland1992adaptation} searches for high-performance legs, which are evaluated on three distinct environments using a high-fidelity simulator which is parallelized across a high-performance compute cluster.  The focus of this paper is to ascertain the viability of this approach.  We pose the hypothesis that {\em evolving selected robot components through high-fidelity simulation automatically generates high-performance morphologies in that environmental niche}. To test this hypothesis we devise two research questions:

\begin{enumerate}
\item Are legs evolved that allow for low-energy traversal of the environment?
\item Do legs evolved in different environments display distinct morphological characteristics?
\end{enumerate}

Initial results show promise; environmental selection generates specialised legs as defined by the hypothesis. Furthermore, the best leg from one environment does not translate well into other environments (in other words, morphologies are specific to the environmental niche considered).  The novelty of this approach is through combining (i) high-fidelity environment simulation, and (ii) a focus on parts of already-useful robots.

The remainder of the paper is organised as follows.  Section~\ref{Related Research} explores related literature in ER and Shape Optimization, Section~\ref{Methodology} describes the approach and implementation, delineates the test problem and describes the simulator, Section~\ref{Genetic Algorithm} details the Genetic Algorithm that searches those representations, and Section~\ref{Results} analyses the results.  Finally, Section~\ref{Conclusions} draws conclusions from the experiments and highlights future work.

\section{Related Research} \label{Related Research}

\subsection{Evolutionary Robotics}

Evolutionary Robotics automatically designs robot morphologies and/or controllers in response to environmental pressure, drawing obvious parallels to the niched development of species in the real world.  Evolutionary motifs have been found to be especially useful in designing robots, for example allowing morphologies to increase their complexity in response to increasingly complex environments~\cite{MorphologicalComplexity}, and abstracting cellular development to 'grow' and 'morph' complex bodies from simple building blocks~\cite{ArtificialMetamorphosis}.  Despite significant leaps in ER, the tasks achievable by evolved robots are far less complex than those expected of a traditionally-built and controlled robot.  Below we describe some common issues, and frame our approach as a way to sidestep some of them.

\subsubsection{Tasks and Environments}
\label{TaskEnv}

Early work evolved robots for locomotion tasks, including including walking, swimming, and jumping~\cite{EvolvingVirtualCreatures}.   Recent examples of tasks for morphologically-evolved robots include locomotion, phototaxis, and environment navigation, however real-world complexity is still generally lacking~\cite{doncieux2011evolutionary}.  Issues include (i) defining a fitness function that captures the desired behaviour, and (ii) evolving control from scratch.  Additionally, tasks may be  complex, requiring modularity or bootstrapping.  We do not optimise for a specific task, instead concentrating on how efficiently the hexapod can stride through a given environment.  The hexapod's pre-existing control stack will automatically exploit the leg when performing any task required of it. 

Owing to the time and cost of repeated physical instantiation, simulation is frequently used as a parallel, cheap, and fast tool to reduce experiment times, e.g.,~\cite{HowToEvolve,howard2014evolving}.  ER experiments require large numbers of iterations of a population of solutions; simulators used in ER therefore tend to emphasise speed over environmental realism and fidelity, being largely devoid of the complexity and high-resolution environmental interactions that real life forms experience and are selected through (see~\cite{doncieux2011evolutionary} for an overview).  As our approach relies on such interactions, we use a powerful simulator that explicitly models high fidelity environmental interactions.  To offset the increased computational burden, the simulator is run on a high-performance computing cluster.

\subsubsection{Representations}

As a robot's controller must be specifically tailored to its body, a popular approach co-evolves control and morphology, producing robots that display creativity and adaptation (see early~\cite{EvolvingVirtualCreatures} and later~\cite{EvolvingRobotMorphologyAndControl} examples).  Representing an entire robot is difficult, and co-evolving a controller exacerbates this issue.  High dimensionality of the search space is a common limiting factor, hinting at a generative representation to create more complex bodies, e.g.~\cite{EvolvingCompleteRobots}.  Generative encodings are typically less evolvable~\cite{tarapore2016different}, and display more complex genotype-to-fitness mappings which must be properly handled.

To simplify the representation, we opt to constrain the morphological search to specific, performance-critical parts of an established morphology.  Given this targeted strategy, and to encourage evolvability, we use a direct encoding.  As much of our morphology is fixed, standard robot control software can be used.

\subsubsection{Reality}

An established goal of ER is to create robots that work in reality, which was demonstrated in 2000~\cite{AutomaticDesign} and now appears increasingly frequently in the literature (e.g.,~\cite{hiller2012automatic}).  This includes a recent trend towards development of automated testbeds that accurately assess the performance of evolved robots, e.g, legged~\cite{ATestbedThatEvolves} and flying~\cite{howard2017platform}.  

Hardware experimentation is slow and expensive, so simulation (as mentioned in Section~\ref{TaskEnv}) is often used.  As ER simulators tend to be fast rather than detailed, and because simulation necessarily abstracts reality to some degree, simulated robots display decreased performance when physically created, which is known as crossing the {\em reality gap}~\cite{jakobi1995noise}.  Mixing simulation and reality is a promising area of research, through e.g. internal models~\cite{bongard2006resilient, cully2015robots}, however it is unknown if model fidelity can match complex real-world environmental conditions.

We cannot feasibly create every single leg, so instead use a highly realistic simulator to attempt to narrow the reality gap.  To further facilitate real-world experimentation, our legs are stored and simulated as 3D-printable files (.OBJ/.STL) for painless instantiation.

\subsection{Shape Optimisation}

Shape optimisation can be divided into two main approaches; gradient-based~\cite{bletzinger2010optimal}, and evolved~\cite{arias2012multiobjective}, with typical targets including free-form architectural design and aerospace structural design.  Of the two approaches, evolution carries two main benefits; (i) it can optimise arbitrary non-convex shapes, and (ii) handles complex multi-modal design spaces and highly non-linear objective functions (e.g.,~\cite{sigmund2011usefulness}).

Designing a robot component bears strong similarities to the evolution of physical artefacts, including jewellery~\cite{AestheticEvolutionary}, sculptures~\cite{CreativeGeneration,UploadAnyObject} 2D art~\cite{Picbreeder}, together with more functional objects including wind turbines~\cite{WindTurbines}.  Typical representations include direct (e.g., Bezier splines~\cite{Cinque1998821}), and generative approaches including Compositional Pattern Producing Networks~\cite{stanley2007compositional} (CPPNs).  We specifically note recent efforts~\cite{7559707} that evolve CPPNs to optimally distribute limited material across a structural joint (similar to our approach, focusing on a small yet critical part of a larger whole).  Evolved joints are shown through Finite Element Analysis (FEA) to be significantly stronger than those produced by other state of the art techniques. 

\subsection{Summary}
Overall, our approach focuses on (i) morphological change based on detailed environmental interactions, (ii) targeted evolution on specific parts of a robot that are involved in environmental interaction, reducing dimensionality, and (iii) evolving components that can be easily printed and used on pre-existing capable robotic platforms.

\section{Experimentation}
\label{Methodology}
Experimentation is designed to assess performance and morphological variations between populations of hexapod legs when evolved in three different environments.  In each environment, a population of 20 legs is randomly generated within respective ranges shown in Section~\ref{REP}, evaluated, and assigned a fitness.  Each population is then optimised for 100 generations.  Each generation comprises (i) creation of 20 new legs via the GA, (ii) generation of an .OBJ file per leg, (iii) simulated evaluation using the .OBJ, and (iv) fitness assignment.  Statistics are saved every generation, including fitness progression and morphological metrics, together with 3D-printable files for each population member.

\subsection{Simulator}
We evaluate legs using the Chrono:Engine simulator\footnote{http://projectchrono.org}, which specialises in high-fidelity environmental modelling, and has previously been used to analyse robotic performance in environment-specific tasks including river fording and traversal of rough terrain for subsequent deployment on real platforms.  It can handle rigid-body, flexible-body, and fluid dynamic environments.  It also considers friction, foot slip, actuator torque limits, and realistic sensor and actuator noise. We model three distinct environments which are likely to be encountered in the real world, and pose different challenges to the robot to encourage morphological specialisation;

\begin{enumerate}
\item A deformable terrain model with the characteristics of soil,
\item A particle-based solid model representing loose gravel, and,
\item A particle-based fluid model that acts like water.
\end{enumerate}

\begin{figure}[thpb]
      \centering
      \includegraphics[scale=0.45, angle=90]{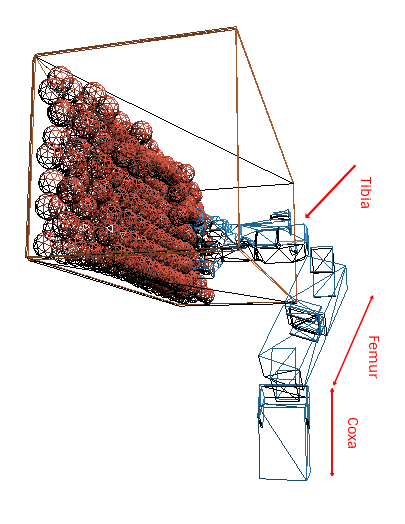}
      \caption{Visualisation of a simulation in the gravel environment. Leg anatomy is labelled.  The leg is commanded to produce a full stride in the environment, from which the required torque is measured and used as a measure of fitness.}
      \label{fluid}
  \end{figure}

Particle-based environments are restricted to a virtual container providing $\approx$7cm of material depth, see Figure~\ref{fluid}.  Importantly, native support is provided to distribute the simulation effort in parallel across high-performance computing nodes; (i) each experimental repeat is run in parallel (ii) all evaluations in a generation are conducted in parallel, and (iii) each individual particle-based evaluation distributes across a compute node.  Despite this, mean times for a full experiment set (10 repeats in an environment) are 7 hours, 2.7 days, and 9.4 days respectively.  Before simulation, legs undergo a brief FEA using conservative force estimates applied to each face of the leg (derived from the simulator), and are rejected if the FEA breaks them such that the top and bottom of the leg are no longer be connected.

\subsection{Target Robot}

\begin{figure}[t!]
      \centering
      \includegraphics[width=8cm]{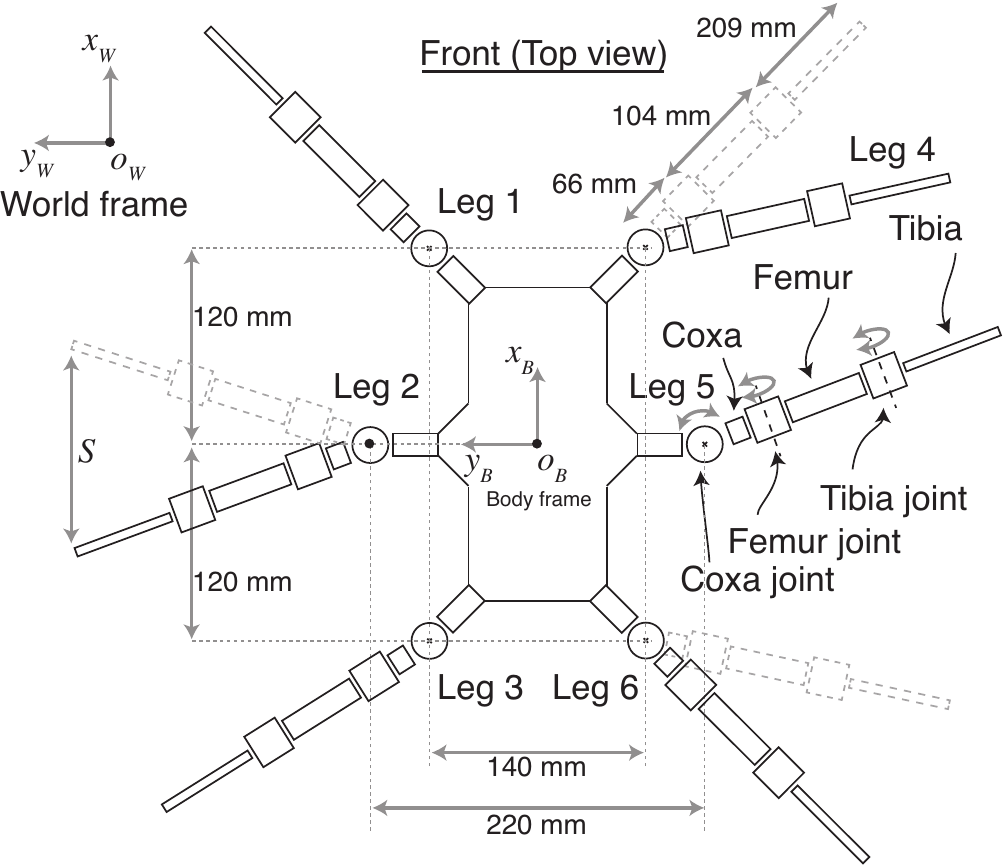}
      \caption{Reference figure showing the hexapod used in the experiments.  We evolve only the leg, as this component directly interacts with the environment.}
      \label{hexapod}
  \end{figure}

Our target robot is an 18 degree-of-freedom hexapod, with dimensions and actuator limits over-viewed in Figure~\ref{hexapod}.  There are currently five of these platforms used by CSIRO for field operations in varied environment types; improving mission performance through morphological optimization of these hexapods is the final goal of this project.  We exploit a mature control stack, developed over the past five years and implemented in ROS, which allows the hexapod to perform autonomous missions.  We focus on morphological adaptation whilst maintaining this capability.  Hexapod bodies do not directly interact with the environment, allowing us to optimise legs only.  Although (co-evolutionary) parameter tuning may be beneficial to fully exploit the morphologies, the current work focuses on morphology only.  Our approach largely preserves the hexapod's morphology, as such the default control stack is assumed to provide sufficient control performance.

\subsection{Leg Encoding}\label{REP}

Hexapod legs are encoded as a set of 5-10 Bezier splines (e.g.,~\cite{Cinque1998821}), and each spline is defined by 3-8 control points (range [0,16] in $x$ and $z$, [0,32] in $y$, $y$ is length), and a thickness of 1-3 voxels.  Each variable-length genome is a collection of Beziers, and each Bezier is a set of control points.  These splines collectively define the occupancy of a voxel grid of resolution$16\times32\times16$\footnote{Heuristically determined. Each voxel represents a 5mm cube in real space}.  If a Bezier intersects a voxel, the voxel is full of material, otherwise it is empty.  

\begin{figure*}[thpb]
  \centering
  \includegraphics[height=8cm]{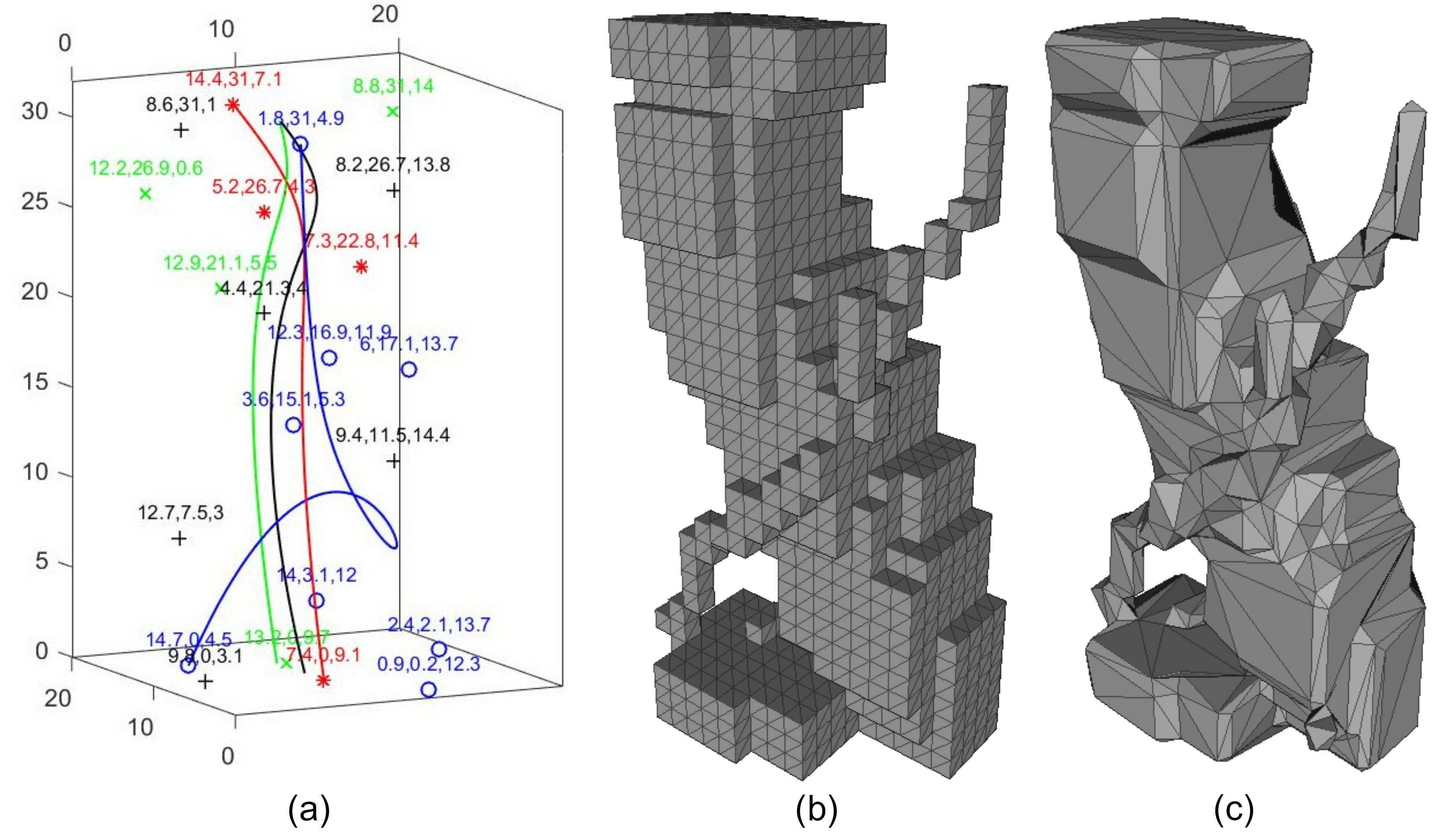}
  \caption{(a) 3D plot of beziers (lines) using control points (circles); Cartesian coordinates with maximum $x$/$z$=16, maximum $y$=32 (b) Voxelised .STL of (different) Bezier splines (c) Post-processed mesh converted into .OBJ}
  \label{mapping}
\end{figure*}

Figure ~\ref{mapping} shows the representation of a leg, and the process of mapping genotype to phenotype.  Beziers are scaled so that each complete leg occupies the full length of the voxel grid, e.g., all legs are the same length --- this ensures that legs are simulated properly.  Future work will consider different leg sizes.

Robot legs are represented in the simulator as an .OBJ file.  Legs are resampled (smoothing the appearance and removing non manifold edges), decimated (reducing the number of faces by $90\%$) for faster evaluation, and rescaled and transformed for instantiation in the simulator using Meshlab\footnote{http://www.meshlab.net/}.  Prior to printing, a generic mounting bracket is added.  The bracket contains holes into which threaded inserts are pushed, allowing the leg to be easily attached to, and detached from, any of our medium-sized hexapods.

\subsection{Fitness evaluation}
During an evaluation, we command the hexapod controller to walk forwards with maximum velocity (1 m/s), which is a commonly-desired behaviour.  The hexapod controller generates foot-tip arcs which the actuators follow through Inverse Kinematics.  The coxa begins rotated $30^{\circ}$ forward and sweeps $60^{\circ}$ backwards; the femur begins elevated by $30^{\circ}$ and throughout the simulation lowers to neutral and raises back up to $30^{\circ}$; finally, the leg follows the inverted path of the femur starting $30^{\circ}$ declined, raising to neutral and then rotating back down $30^{\circ}$.  Actuators are modelled to obey realistic range limits and output torques taken from the actual devices.  An evaluation lasts for a single hexapod stride, which is 3000 1ms simulation steps. Figure~\ref{fluid} shows a snapshot of an evaluation. 

Fitness is proportional to the torque required to move the leg through the environment; this effectively minimises cost of transport, which is a typical feature of locomotion strategies in natural creatures.  We assume that the hexapod walks forward, precluding alternative behaviours that are beneficial, yet require higher torques (such as swimming using paddle-like appendages).

A second term penalises excessive material usage and discourages superfluous appendages.  The fitness function can be seen in \cref{eq:fitness}, where $\tau$ is the total cumulative torque for the three leg actuators, $n_{steps}$ is the number of simulation steps and $\delta$ is the percentage of occupied voxels in a given leg.  We seek legs that minimise this function.    Note that our robot is already fully controllable, and can be commanded to reach any speed through its gait engine, which we leave unmodified.  Torque relates to minimizing energy loss when interacting with the environment for a given stride length and speed.

\begin{equation}
    f = \frac{\tau}{n_{steps}}+\bigg(\frac{ \tau}{n_{steps}}\times \frac{\delta}{5}\bigg)
    \label{eq:fitness}
\end{equation}

\section{Evolutionary Algorithm} \label{Genetic Algorithm}

The initial population of legs are randomly initialized from a uniform distribution within allowed ranges.  The population is evaluated in the chosen environment and fitness values are assigned.  Each generation, twenty new legs are generated.  For each, two parents are selected via tournament selection with a tournament size of 4.  Parents donate to the child using two-point crossover,  with the two crossover points random-uniformly chosen within the range of the shortest parent genotype.  Mutation varies each control point using a Gaussian with a mean of 0 and a Standard Deviation $\sigma$ of $10\%$ of the range for that Cartesian co-ordinate of each control point. 

We employ a number of structural mutations; each of these probabilities is drawn from a uniform distribution in range [0,1].  With P=0.2, a random Bezier's thickness is randomly reinitialized within the allowed range. With the same probability, a random Bezier is selected and a control point either added (P=0.5) or removed from it.  With P=0.1, a new Bezier with a random number of control points is either added (P=0.5) or removed from the leg.  Each newly-created leg is then evaluated and assigned a fitness.  The combined parent and child population is then fitness ranked, and the top half taken as the new population for the subsequent generation.

\section{Results} \label{Results}

Ten repeats of each environment are run, producing the statistics analysed herein, and permitting statistical comparisons using the Mann-Whitney U-Test\footnote{Which does not require normally-distributed samples.}, with significance at P$<$0.05.  The 30 experiments were run simultaneously on a high performance compute cluster using a total of 600 cores, and condensed approximately 100 days worth of evaluations down to 9.4 days.

\subsection{Performance}

\begin{figure}[t]
\begin{center}

\subfloat[] { \includegraphics[width=8.5cm]{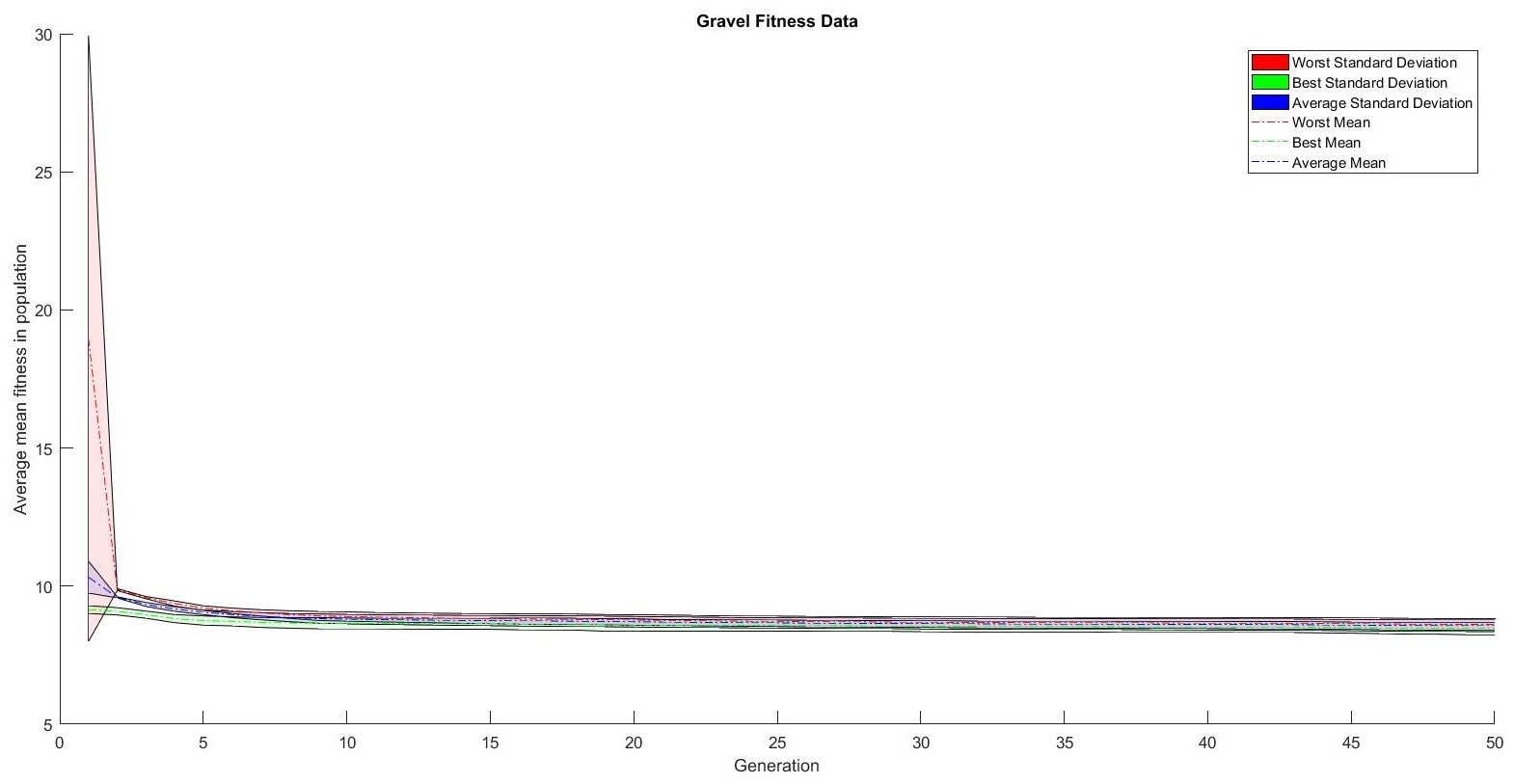}} \\
\subfloat[] { \includegraphics[width=8.5cm]{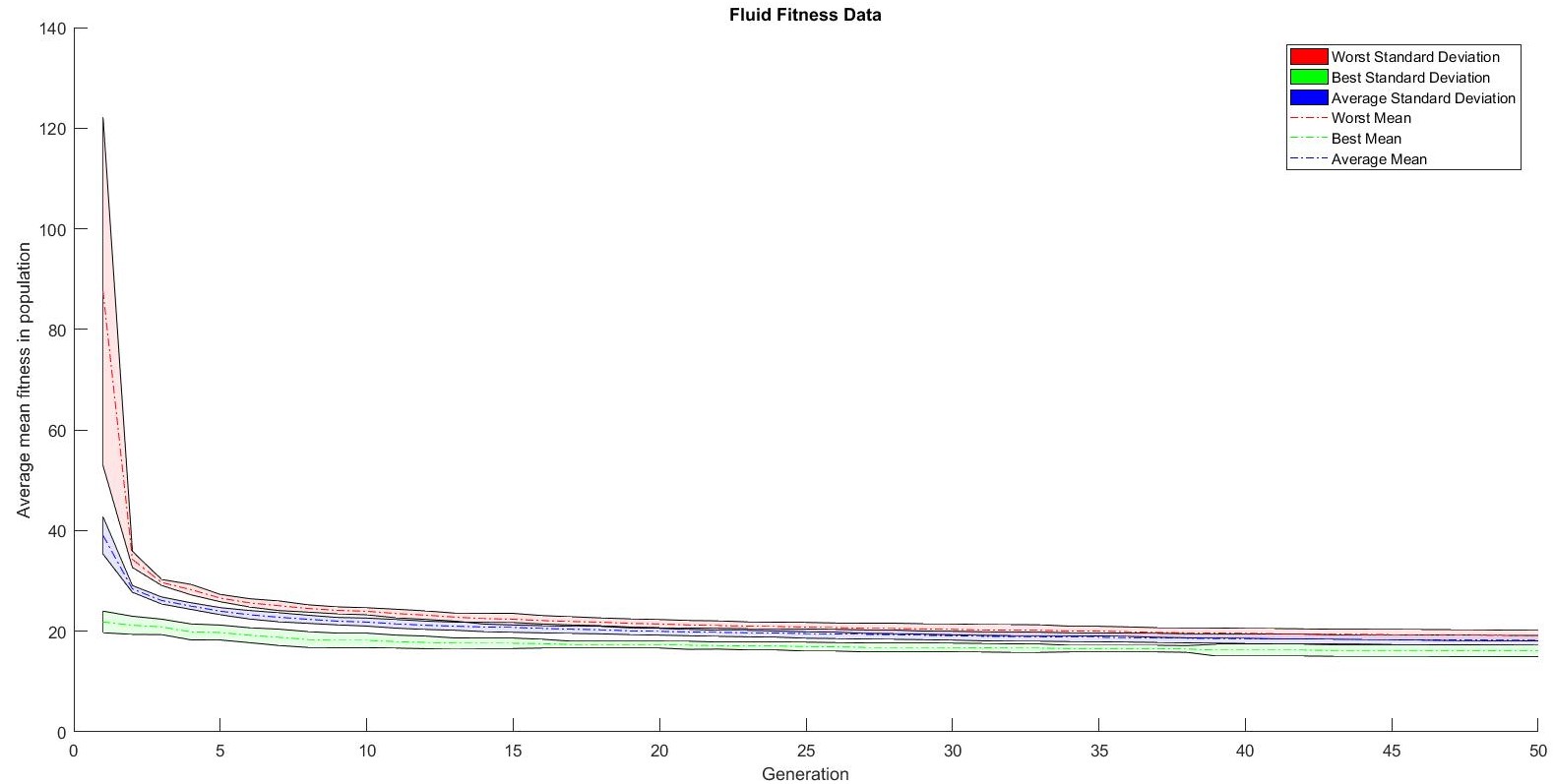}} \\
\subfloat[] { \includegraphics[width=8.5cm]{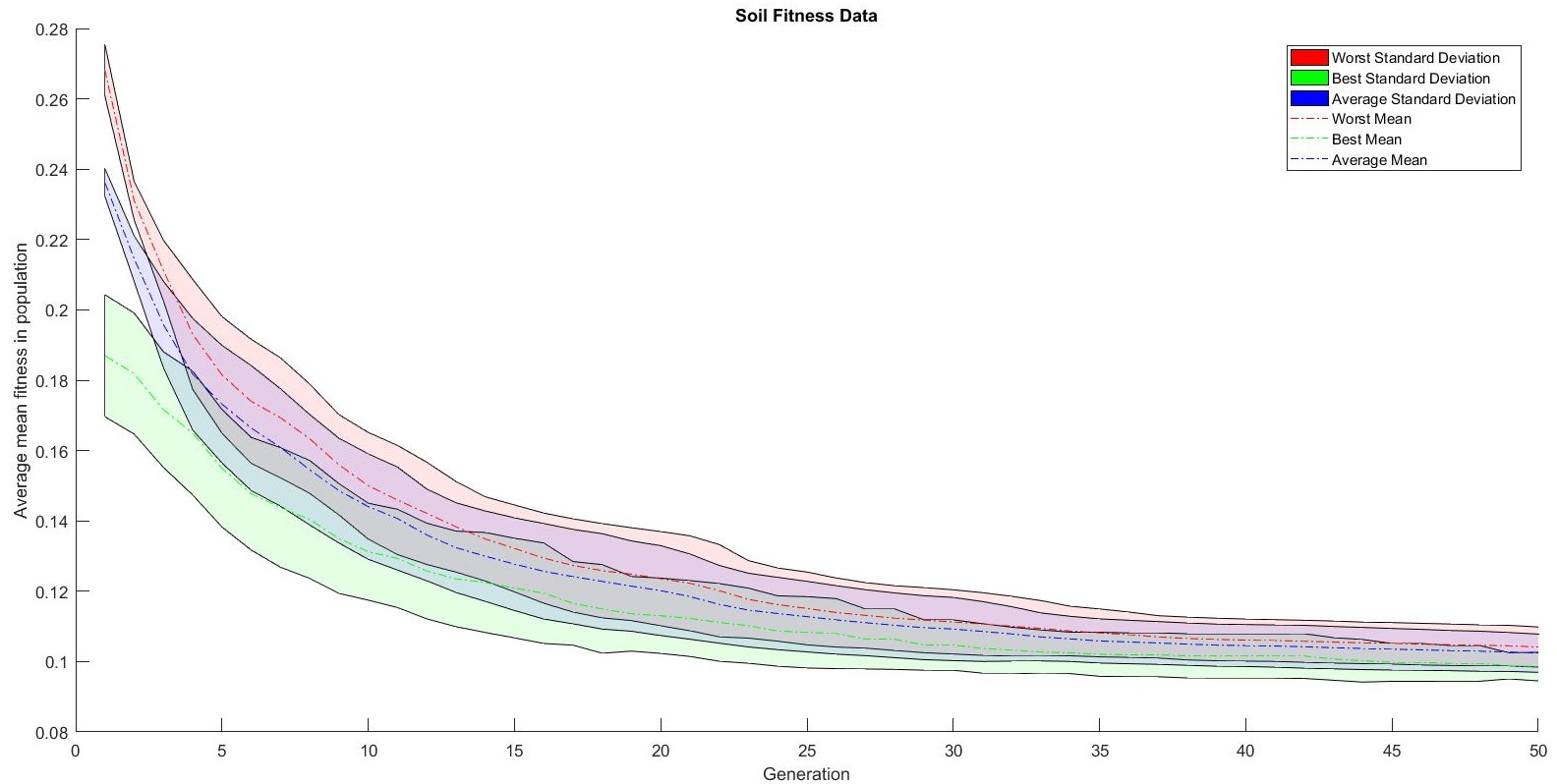}} 
\end{center}
\caption{Showing fitness progression through the first 50 generations, for (a) gravel, (b) fluid, and (c) soil environments.  Green = best fitness, blue = mean fitness, red = worst fitness.  Shaded areas denote standard error.}
\label{Fitness}
\end{figure}

Fitness graphs hint at the comparative difficulty of the environments; gravel (Figure~\ref{Fitness} (a)) converges within 20 generations, followed by fluid (Figure~\ref{Fitness} (b)) at $\approx$40 generations, and lastly by soil (Figure~\ref{Fitness} (c)) within 80 generations.  Fitness convergence is evidenced within 100 generations in all repeats for each environment.  Converged best, average, and worst fitness values are statistically different between the environments. 

Notable fitness improvement occurs in all cases.  Average best fitness for gravel legs commences at 8.7, quickly lowering to a final value of 8.1.  Similar trends are evidenced across gravel (average mean fitness starts at 10.8 and ends at 9.0, average worst fitness starts at 18.8 and ends at 9.2).  The fitness profile for fluid closely follows that of gravel; best fitness starts at 21.5 and ends at 19.1, mean fitness quickly decreases from 39.4 to 23.1, and worst fitness drops from 123.1 to 24.2 within the first 20 generations.  Soil shows markedly different fitness profiles when compared to the other two environments, dropping more gradually across the generations.

To show environmental specialisation, we next take the best leg per experimental repeat per environment, and evaluate them in the other two environments.  Table~\ref{compTable} shows that the best leg in each environment is the one generated in that environment.  Interestingly, the slimmer legs generated in the soil environment perform poorly in the other environments.  For gravel, the contact point is small and singular, so the leg penetrates into the gravel and results in increased torques required to drag the leg.  In fluid, the superfluous appendages that usually have no detrimental effect result in increased torques as they pass through the fluid.  This partially confirms our theory that flat surfaces are an advantage for fluid.

\begin{table}[h]
\caption{Showing the average fitness attained by the best leg per experimental repeat (rows) when evaluated in the other environments (columns).}
\label{compTable}
\begin{center}
\begin{tabular}{|c|c|c|c| }
\hline
 & Fluid & Gravel & Soil\\
\hline
Fluid & 19.1 & 8.6355 & 0.1537\\
\hline
Gravel & 58.5300 & 8.0856 & 0.1404\\
\hline
Soil & 86.9735 & 117.0014 & 0.0904\\
\hline
\end{tabular}
\end{center}
\end{table}

\subsection{Morphology}

\begin{figure*}[thpb]
  \centering
  \includegraphics[width=\textwidth,height=9cm]{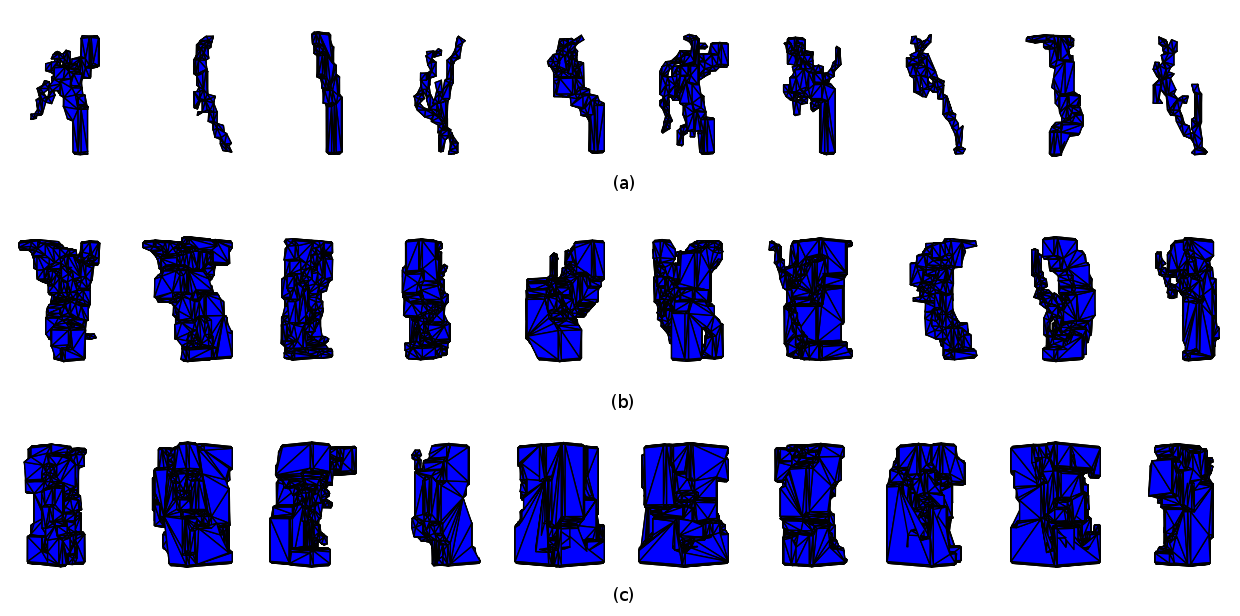}
  \caption{(a) Legs evolved for deformable terrain model with the characteristics of soil, (b) legs evolved for a particle-based solid model representing loose gravel, (c) legs evolved for a particle-based fluid model that acts like water.}
  \label{optimalLegs}
\end{figure*}

The most optimal legs from all 30 experiments are pictured together in Figure ~\ref{optimalLegs}, grouped by environment type.  We note interesting morphological differences between the three, including:

{\bf Soil} legs being generally skinnier than the other two; as the leg cannot 'fall through' the soil, the second term in the fitness function has more influence on the final morphology.

{\bf Gravel} legs having more bulk than soil legs, so as not to fall through the environment and require additional torque resulting from dragging a partially 'submerged' leg through the gravel.
 
{\bf Fluid} legs being perhaps the most surprising, as they are similar to gravel legs.  It was unclear why larger legs were generated, as reducing torque through fluid intuitively suggests a streamlined, thin design.  Post-hoc analysis reveals that, regardless of their morphology, legs sunk to the bottom of the environment container (as expected).  However, completing the movement through the fluid incurred large torque penalties regardless of morphology that were difficult to optimize through; it is possible that legs converged prematurely on an initial 'bulky' design or local minima --- see the best fitness plot in Figure~\ref{Fitness}(b).  Conversely, large, flat surfaces in certain configurations may be easier than imagined for the actuators to drag through the environment, and the actual torque differences between thinner and blockier legs may be less than is appreciable through our distal viewpoint.

\begin{figure}[t]
\begin{center}
\subfloat[] { \includegraphics[width=4cm]{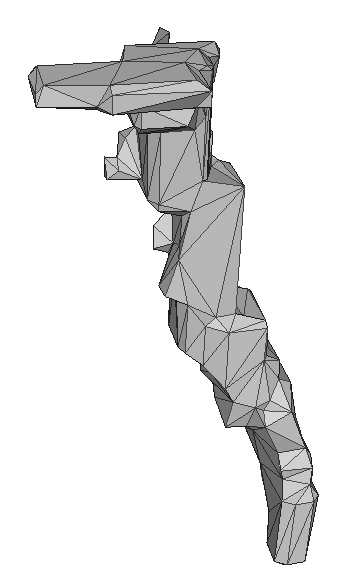}} \\
\subfloat[] { \includegraphics[width=4cm]{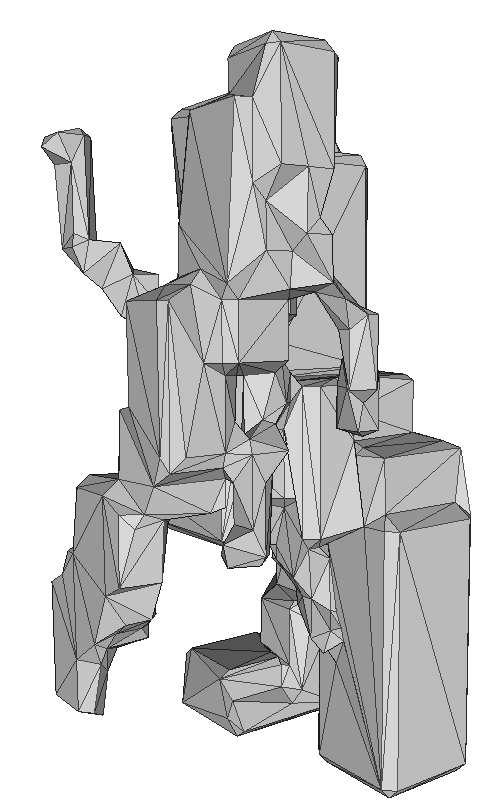}}
\end{center}
\caption{Representative legs evolved in (a) soil and (b) gravel environments.  Note larger legs in (b), allowing multiple contact points to the environment and preventing the leg becoming partially submerged, resulting in better fitness.}
\label{legs}
\end{figure}

Given this result, we shift our focus towards soil and gravel legs (representative legs seen in greater detail in Figure~\ref{legs}).  There are profound morphological differences between the legs, which is confirmed through counts of occupied voxels (mean 425 for soil, 2071 for gravel, averaged over the best leg per environment).  We postulate that larger legs are evolved for gravel as they more easily facilitate the development of a larger footprint, creating multiple contact points and allowing the leg to better sit atop the gravel which reduces the required torque.  Soil legs, optimized for solid ground, favour a single contact point.  Decreased torque pressure from the environment results in the voxel-reducing fitness term having a more prominent effect on final morphology, encouraging legs to 'overlap' their occupancies for better fitness scores.

In general, designs have a large contact area with the mounting bracket (which would be added to the top layer of the leg); this improves the likelihood of the leg surviving the FEA test.  Interestingly, we note that some high-performance designs would likely fall inside of an engineers' design space, whereas others are more unconventional in appearance, and more likely uniquely discoverable through evolution.  We also note seemingly superfluous artefacts, most prominently in legs 1, 6, and 10 of Figure~\ref{optimalLegs}(a), and to a lesser extend in legs 1, 9, and 10 Figure~\ref{optimalLegs}(b).  While adding to the number of occupied voxels, and being far away from any environmental contact, such artefacts are frequently the terminal parts of splines that add useful structure or increase surface contact.  The algorithm struggled to fully remove these superfluous artefacts, which suggests an alternate representation may be preferred.

To further study the morphological aspects of the evolved legs, we compare the best legs evolved in each run, based on the percentage of similarly-occupied voxels in the grid. legs evolved in the same environment are most similar, with a maximum of $\approx$92\% correlation for soil-soil.  Gravel and liquid legs are more dispersed around the design space, indicating a more multi-modal problem, and correlating with the higher best fitness values for those environments.

Multiresolution Reeb Graphs~\cite{ReebGraph} provide an established method to compare morphological similarity between the .OBJ files of the optimised legs. Table~\ref{corrReeb} shows the highest percentages are achieved between legs from the same environment, reinforcing the notion that legs from the same environment are morphologically similar.  Note that high numbers in general are expected as we compare objects from the same class (legs).

\begin{table}[]
\centering
\caption{Reeb Graph summary, showing percentage similarity in the phenotypes (smoothed and decimated .OBJ files) between the best legs per experimental repeat, per environment. }
\label{corrReeb}
\begin{tabular}{|l|l|c|}
\hline
\multicolumn{2}{|l|}{Correlation between:} & Average Percentage \\ \hline
Soil                 & Soil                & 78.2      \\ \hline
Soil                 & Gravel             & 73.1      \\ \hline
Soil                 & Fluid               & 69.5      \\ \hline
Gravel              & Gravel             & 77.8      \\ \hline
Gravel              & Fluid               & 74.3      \\ \hline
Fluid                & Fluid               & 76.9      \\ \hline
\end{tabular}
\end{table}

\section{Conclusions} \label{Conclusions}

We investigated the use of a high-resolution simulator to generate environment-specific legs for a hexapod robot. The simulated legs are easily 3D printable, so combined with a basic fitness function of reducing energy when walking, can provide power-saving components for our collection of real hexapods when completing a wide variety of missions in myriad environments.  The hexapod preserves its body morphology and mature, well-supported control software, and as such can complete a wide variety of useful missions with the benefit of environmentally specialised morphologies that are targeted in areas most likely to interact with the environment.

The next step is to collect extensive results from field tests when using these legs in real operations to prove the real-world applicability of this approach.  However, in this initial research we have proven our hypothesis; legs are shown to maximise their performance in each environment.  Morphological analysis supports the notion that different environments create distinctly-shaped legs (both visually and through mathematical analysis), and furthermore we show that legs evolved in one environment do not transfer well to other environments.  Legs are therefore morphologically specialised to save energy.  Proving this hypothesis motivates further exploration into this type of targeted environmental optimisation.

Numerous extensions are possible, including;

\begin{itemize}
\item Real-world validation,
\item Injecting real-world test results back into the simulator,
\item Exploring generative codings and diversity selection, and,
\item Removing restrictions on the controller, so that the stride is mutable and the hexapod may learn to 'skate' over ice, for example.  This will likely take the form of constrained co-evolution.
\end{itemize}

We also note that we optimise forward motion only; turning and other manoeuvres are not currently considered. As a proof of concept, however, we see encouraging signs that continuation of this line of research will be fruitful in melding engineered and evolved approaches for sophisticated behaviours with morphological environmental specialisation.

\bibliographystyle{ACM-Reference-Format}
\bibliography{bibliography-conf} 

\end{document}